\pdfoutput=1

\documentclass[11pt]{article}

\usepackage[hyperref]{emnlp2021}

\usepackage{times}
\usepackage{latexsym}
\usepackage{graphicx}
\usepackage{booktabs}
\usepackage{amsmath}
\usepackage[T1]{fontenc}

\usepackage[utf8]{inputenc}

\usepackage{microtype}

\title{Towards Expressive Communication with Internet Memes: A New Multimodal Conversation Dataset and Benchmark
}

\author{    Zhengcong Fei$^1$$^2$, Zekang Li$^1$$^2$, Jinchao Zhang$^3$\thanks{\ \ Joint work with Pattern Recognition Center, WeChat AI, Tencent Inc. Jinchao Zhang is the corresponding author. Work was done when Zhengcong Fei and Zekang Li were intern at WeChat AI.}, Yang Feng$^1$$^2$, \bf{Jie Zhou}$^3$ \\
  $^{1}$ Key Laboratory of Intelligent Information Processing \\
  Institute of Computing Technology, Chinese Academy of Sciences (ICT/CAS) \\
  $^{2}$ University of Chinese Academy of Sciences \\
  $^{3}$ Pattern Recognition Center, WeChat AI, Tencent Inc, China \\
  {\tt \{\href{mailto:lizekang19g@ict.ac.cn}{lizekang19g},\href{mailto:feizhengcong@ict.ac.cn}{feizhengcong},\href{mailto:fengyang@ict.ac.cn}{fengyang}\}@ict.ac.cn} \\
  {\tt \{\href{mailto:dayerzhang@tencent.com}{dayerzhang},\href{mailto:withtomzhou@tencent.com}{withtomzhou}\}@tencent.com}
  \\}

\begin{document}
\maketitle
\begin{abstract}
As a kind of new expression elements, Internet memes are popular and extensively used in online chatting scenarios since they manage to make  dialogues vivid, moving, and interesting. However, most current dialogue researches focus on text-only dialogue tasks. In this paper, we propose a new task named as \textbf{M}eme incorporated \textbf{O}pen-domain \textbf{D}ialogue (MOD). Compared to previous dialogue tasks, MOD is much more challenging since it requires the model to understand the multimodal elements as well as the emotions behind them. To facilitate the MOD research, we construct a large-scale open-domain multimodal dialogue dataset incorporating abundant Internet memes into utterances. The dataset consists of $\sim$45K Chinese conversations with $\sim$606K utterances. Each conversation contains about $13$ utterances with about $4$ Internet memes on average and each utterance equipped with an Internet meme is annotated with the corresponding emotion. In addition, we present a simple and effective method, which utilizes a unified generation network to solve the MOD task. Experimental results demonstrate that our method trained on the proposed corpus is able to achieve expressive communication including texts and memes. The corpus and models have been publicly available at \url{https://github.com/lizekang/DSTC10-MOD}.
\end{abstract}

\section{Introduction}
Internet memes have become one of the most important approaches for expression and emotions in social media \cite{DBLP:journals/pacmhci/WangLGKL19}. 
Compared with text-only communication, dialogues become more expressive and vivid when Internet memes are incorporated.
Meanwhile, there are many implicit and strong emotions delivered by Internet memes.
Resultingly, the use of Internet memes in online chats has become increasingly popular \cite{DBLP:conf/hci/Chen20,DBLP:journals/ipm/BeskowKC20}.
Even though there is an increasing interest in chatbots that can converse with humans using multiple modalities, incorporating contextualized Internet memes into multi-turn open-domain dialogues under diverse situations is still far from satisfactory. 

\begin{figure}
	\begin{center}
		\includegraphics[width=0.95\columnwidth]{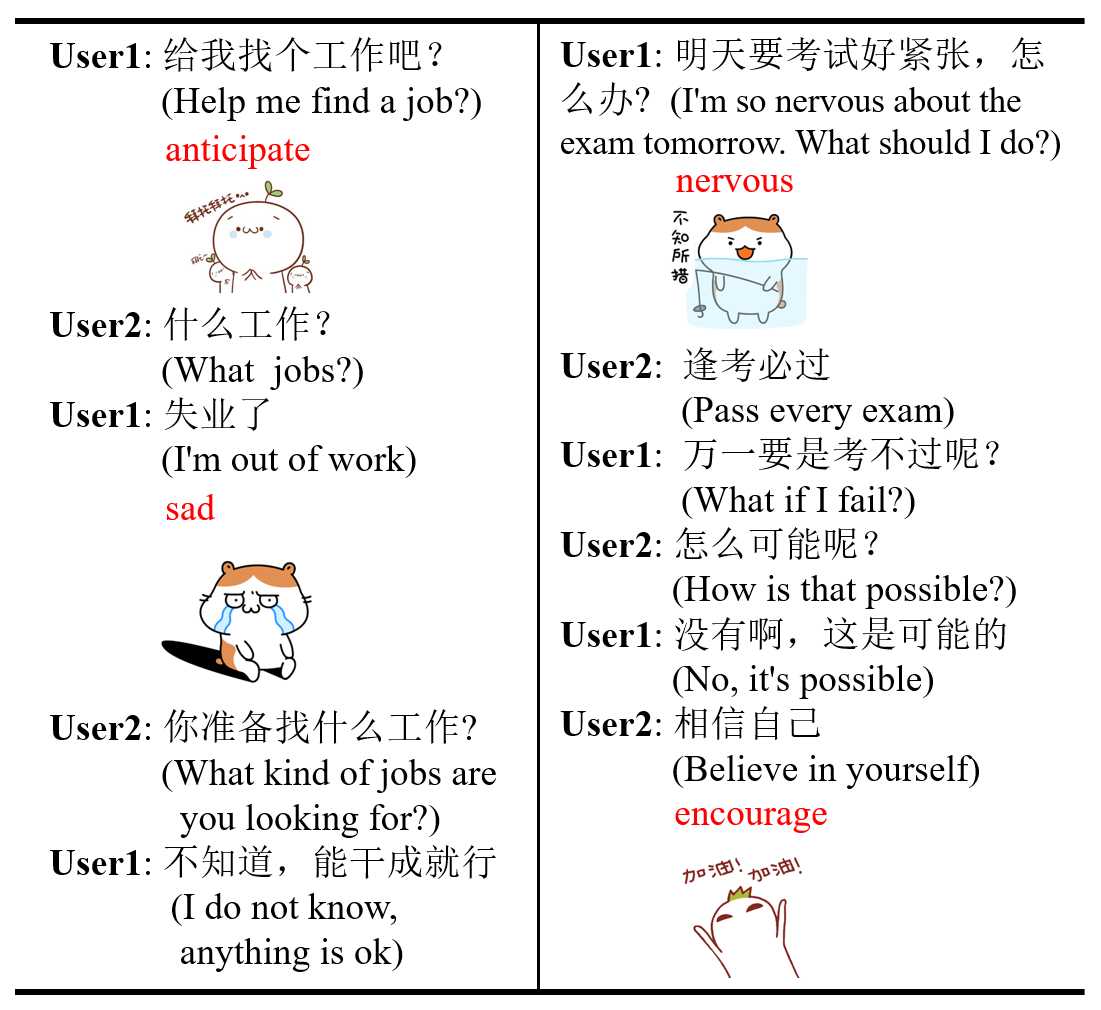}
	\end{center}
	\caption{Illustrations of meme incorporated open-domain dialogues. Both history and response can be in the form of text-only, meme-only, or a combination of both. Corresponding emotion is annotated for each used Internet meme in red. }
	\label{fig:1}
\end{figure}

A variety of related works have been proposed to explore the multimodal information in dialogues.
The first group of work is emoji prediction \cite{DBLP:journals/corr/XieLYS16,DBLP:conf/eacl/SaggionBB17,DBLP:conf/naacl/BarbieriBRS18} according to the chatting context. 
The emojis are limited in variety and of small size, while memes are more expressive and constantly evolving.   
The second group is retailing multimodal dialogue, which is task-oriented and the utilized images are the products on sale \cite{DBLP:conf/mm/LiaoM0HC18,DBLP:conf/mm/NieWHWT19,DBLP:conf/sigir/CuiWSHXN19}. 
The models developed for this task often focus on specific aspects and then generating the correct answers to questions, which fails to model those important aspects for dialogue such as emotion analysis. 
Another group of research works investigate visually grounded dialogue \cite{DBLP:conf/cvpr/DasKGSYMPB17,haber-etal-2019-photobook}, which uses natural language to talk about visual content given in advance. However, the mode of conversation there is limited to only text \cite{DBLP:conf/acl/LeSCH19}.

As a step towards expressive open-domain conversational AI, 
we introduce a new task -- Meme incorporated Open-domain Dialogue (MOD), along with a large-scale Chinese multimodal conversation dataset, which facilitates the multimodal conversation modeling and emotion analysis in multi-turn dialogues. 
Specifically, provided with a multimodal dialogue context, the MOD task aims to generate a vivid response in text-only, meme-only, or mixed information, which can be considered a general paradigm compared with conventional dialogue tasks.
Our dataset consists of $\sim$45K human-human open-domain Chinese conversations between two participants.  Each meme in the conversation is annotated with the corresponding emotion, which can be used as the supervision information for emotion analysis modeling.  In particular, we provide a hard version of testing set which contains memes not appeared during training to evaluate the generalization ability of model comprehensively. 
Examples of conversations in our dataset involving both text and Internet memes are shown in Figure \ref{fig:1}.

To showcase how the new dataset may be exploited, we propose a simple and effective model, which utilizes a unified network to generate multimodal responses. 
Practically, we pool all sub-tasks like text generation and Internet meme prediction into a general sequence generation procedure, and solve them with a language model architecture. It avoids designing complex systems for individual components separately, and all resulting sub-tasks can be covered simultaneously.
Experimental results highlight that our proposed model trained on the new corpus can create reasonable combined responses and show promising to develop better models.  It is our hope that the introduction of this task will spark a new interest in multimodal open-domain conversation modeling. 
The main contributions of this paper are summarized as follows:
\begin{itemize}
    \item We introduce a new task where a dialogue system is required to incorporate lively Internet memes into open-domain conversations towards expressive communication.
    \item We build a large-scale Chinese dialogue dataset, which contains abundant Internet memes and emotion annotations. The corpus can empower the researches of not only MOD generation but also emotion modeling.
    \item To further illustrate the dataset's potential, we propose a simple and effective model for reference and conduct extensive experiments. Results show that our method can generate rational multimodal responses, while there is still much room for further research. 
\end{itemize}

\section{Related Work}

\subsection{Internet Meme in Dialogue}
Multiple expression forms, including emoji, sticker, and meme, have become prevalent along with the development of online chats. 
Among them, the meme is a type of content that features a visual format of images, GIF, or short videos, which can inject humor into conversations and create an emotional context \cite{DBLP:journals/ejis/PoseyLRE10}.
Compared with emoji which is restricted at a fixed size, 
\cite{DBLP:conf/naacl/BarbieriBRS18,DBLP:conf/eacl/SaggionBB17}, Internet meme is more expressive and of a great variety. 
One similar work to ours is sticker recommendation \cite{DBLP:conf/ism/JesusCBGCM19,DBLP:conf/aaai/LaddhaHMPN20,DBLP:conf/www/GaoCLL0020}, where suitable stickers are retrieved to match the text-only dialogue history, which can be regarded as a special case of MOD. Besides, the tasks of meme retrieval  \cite{DBLP:conf/icde/MiloSY19,DBLP:journals/corr/abs-2002-01462,DBLP:conf/semeval/SharmaBSSDCPG20}, detecting the hate speech in memes and clustering memes according to events \cite{DBLP:conf/evalita/MilianiGRAL20,DBLP:conf/nips/KielaFMGSRT20} are proposed to help the Internet meme modeling.
In this work, we focus on a more challenging situation, generating Internet meme merged utterances to make conversations more vivid and engaging.

\subsection{Multimodal Dialogue Datasets} 
More researches on dialogue systems have shifted towards integrating more modalities, such as image, audio, and video, to build the informative conversation interaction. The datasets reported in \cite{DBLP:conf/cvpr/DasKGSYMPB17,DBLP:conf/ijcnlp/MostafazadehBDG17} make contributions to bridge the gap between vision and language. 
In the VisDial \cite{DBLP:conf/cvpr/DasKGSYMPB17,DBLP:conf/emnlp/KamezawaNSMN20}, a system is required to answer the questions about an input image given the dialogue history. 
And the AVSD dataset \cite{DBLP:conf/cvpr/AlAmriCD0CEBMHA19} has been used for response generation with audio and visual features \cite{DBLP:journals/corr/abs-2002-00163}. 
But in our dataset, the Internet memes are randomly distributed in conversations rather than as background knowledge.
The Multimodal Dialogue (MMD) dataset \cite{DBLP:conf/aaai/SahaKS18}, which includes  on the fashion domain with the information from both images and texts, has further facilitated the researches on the multimodal dialogues. 
While datasets resulting from the above tasks provide opportunities to explore multimodal dialogues, they are more concerned about question answering and limited by task-oriented scenarios.
On the contrary, the MOD includes Internet memes usage in natural conversations and additional emotion modeling.

\section{Task and Data}
\subsection{Meme incorporated Open-domain Dialogue (MOD) task}

Provided with the dialogue history consisting of utterances filled with Internet memes, the dialogue system aims to build an interesting response in the form of text-only, meme-only, or a mixed category of both.
Formally, we use \textbf{U}$_N = [u_1, \ldots, u_N]$ to denote a $N$ turns of Internet meme incorporated dialogue, where utterance $u_i=\langle{S}_i, m_i \rangle$ includes the text context ${S}_i$ and Internet meme $m_i$ pair. 
Note that $m_i$=None denotes there is no Internet meme incorporated in $i$-th utterance and ${S}_i=\{\text{[bos], [eos]}\}$ denotes that no text is generated as the response. Therefore, the goal of MOD is to predict the target response $\hat{u}_{i}$ for the given dialogue history $\textbf{U}_{i-1}$ as: 
\begin{equation}
    \hat{u}_i = \mathop{\arg\max}_{\langle S_i, m_i \rangle} p( \langle S_i, m_i \rangle |\textbf{U}_{i-1})
\end{equation}

We further split the current scope of MOD into the following three consecutive sub-steps: 
(1) \textbf{Text Response Modeling}: given the multimodal history context $\textbf{U}_{i-1}$, the task aims to generate a coherent and natural text response ${S}_i$. (2) \textbf{Meme Usage Prediction}: given a multimodal context of several dialogue turns $\textbf{U}_{i-1}$ and generated text response ${S}_i$, the task here is to decide whether to involve an Internet meme into response, which can be considered as a binary classification. 
 (3) \textbf{Meme Retrieval}: given a multimodal historical context $\textbf{U}_{i-1}$ and generated text response ${S}_i$, the goal here is to select a suitable meme $m_i$ as feedback. 

\subsection{Data Collection}

In this section, we describe the construction of our proposed multi-turn meme-incorporated dialogue dataset in detail.

\paragraph{Step 1: Pre-processing.} 

For Internet meme sets,  the meme candidates are firstly collected from the Internet and then chosen carefully by annotators to maintain good quality. In addition, if textual information appears in the selected Internet meme content, we will also annotate it manually. 
To avoid the model only utilizing the textual information and ignoring visual features, we control the proportion of memes without appeared texts in the final set to 40\%.
Meanwhile, to avoid multiple appropriate memes being selected under one dialogue condition, we filter out the memes with highly similar or duplicate semantic content.
Finally, we obtain a total of 307 Internet memes for the subsequent data annotating process. 
To facilitate the arrangement and annotating process, 
the Internet meme set is further split into four groups: \emph{atmosphere adjustment, basic expression, basic emotion,} and \emph{common semantics}, respectively.

For the initial conversation set, considering that the open-domain Internet meme is too scarce in scale, it is costly and time-consuming to collect multi-turn conversations from scratch. Thus, our annotation is based on an existing large-scale Chinese dialogue dataset with its large version \cite{DBLP:conf/nlpcc/WangKZHJZH20}. To make each chatting session contain rich information, we remove the dialogues which have less than $10$ utterances.

\paragraph{Step 2: Internet meme incorporated response construction.} 

The annotators, who are well-educated and familiar with dialogue research, are tasked to take two operations using the prepared Internet meme candidates: use one most suitable Internet meme to replace part of the text conversation or insert an Internet meme into the utterance to enhance the emotion of the current dialogues. In particular, we also ask annotators to label the emotional states when utilizing the current Internet memes.  
The annotators are specially instructed based on the following criteria: ($i$) behave naturally, and the meme usage is in line with real daily chats,
($ii$) the number of different Internet memes in the dataset is kept balanced to avoid meaningless gatherings and biased data. 
Those dialogues without any labeled Internet memes will be abandoned in the later data processing stage.
Note that different from previous works, our annotation procedure is conducted posteriorly so that it will not interfere with human conversations, \emph{e.g.}, prompting them to an overused Internet meme.

\begin{table}[t]
\small
	\begin{center}
		\setlength{\tabcolsep}{2.5mm}{
			\begin{tabular}{l|r}
				\toprule
				\textbf{Dataset Statistics} & \textbf{size}\\
				\midrule
				\# dialogues (chat sessions) &45,174\\
				\# utterances &606,014\\
				\# tokens &5,339\\ 
				\# Internet memes &307\\\midrule
				Avg. \# of utterances in a dialogue &13.42\\
				Avg. \# of Internet memes in a dialogue &4.06\\
				Avg. \# of tokens in an utterance &11.46\\
				\bottomrule
		\end{tabular}}
		{\caption{Statistics of the MOD dataset.}
			\label{tab:1}}
	\end{center}
\end{table}

\begin{table}[t]
\small
	\begin{center}
		\setlength{\tabcolsep}{1.mm}{
			\begin{tabular}{l|rrrr}
				\toprule
				&\textbf{Train} & \textbf{Valid} & \textbf{Easy test}&\textbf{Hard test}\\
				\midrule
				\# dialogues&41,644&1,000&1,000&1,530\\
				\# utterances&558,181&13,666&13,999&20,358\\
				\# tokens&5,249&2,724&2,782&3,166\\
				 \# Internet memes&274&274&274&307\\
				\bottomrule
		\end{tabular}}
		{\caption{The split statistics of the MOD dataset.}
			\label{tab:2}}
	\end{center}
\end{table}

\begin{figure}[t]
	\begin{center}
		\includegraphics[width=1\columnwidth]{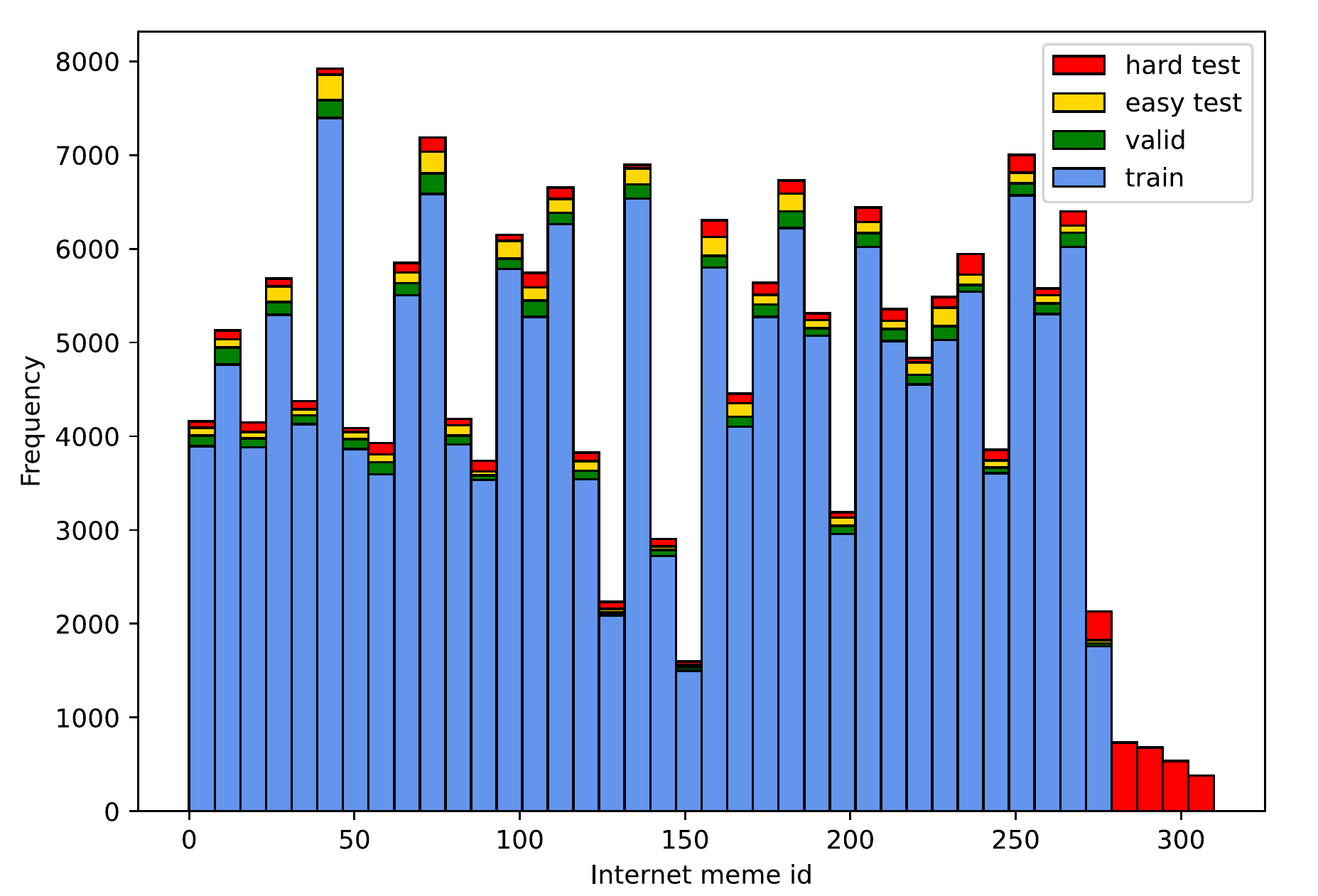}
	\end{center}
	\caption{Internet meme frequency in the dataset. The meme usage balances without significant bias. Meme ids greater than 274 only occur in hard test set.}
	\label{fig:2}
\end{figure}


\begin{figure}[t]
	\begin{center}
		\includegraphics[width=0.9\columnwidth]{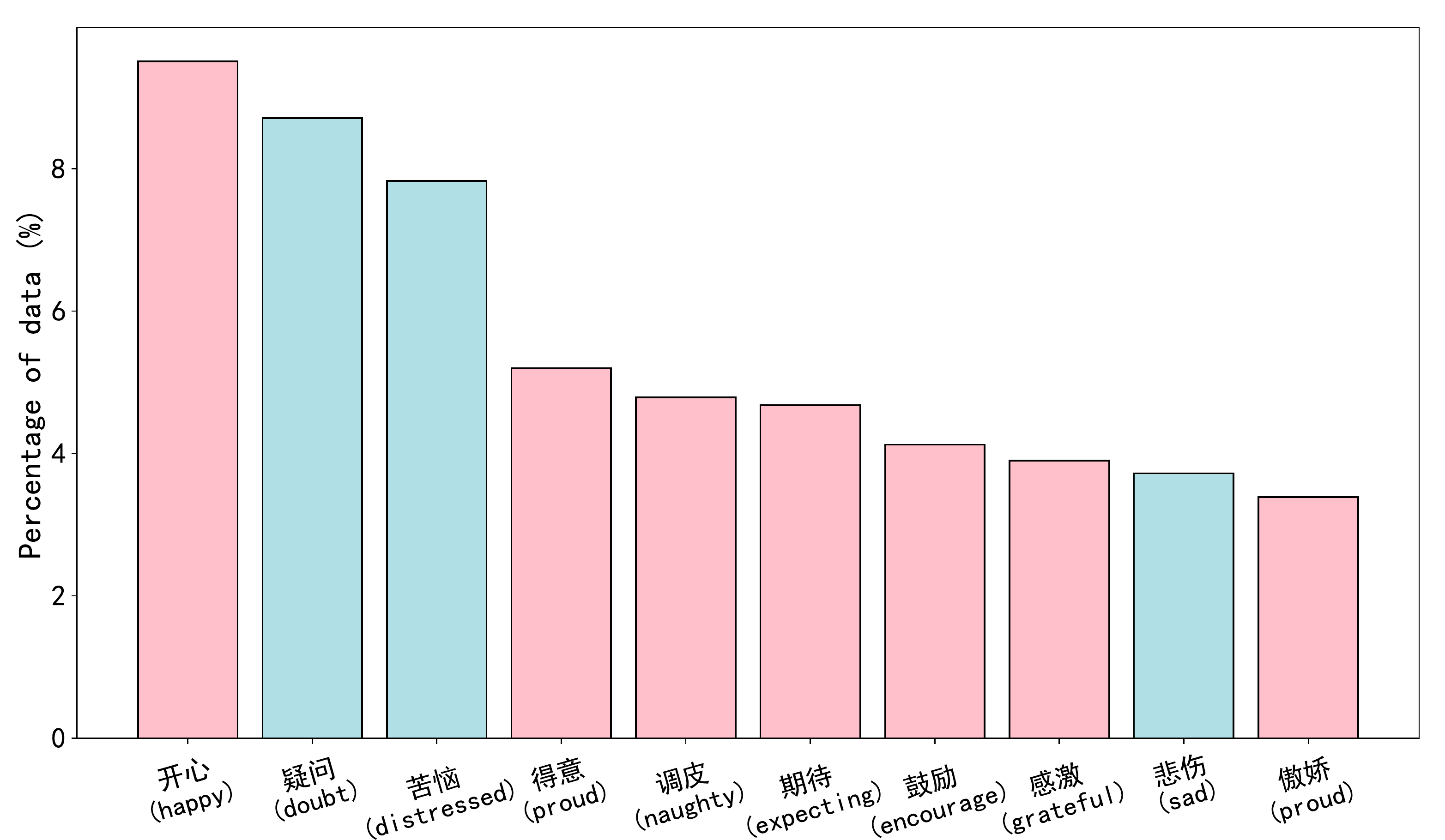}
	\end{center}
	\caption{Histogram of top-10 annotated emotions when memes are used. Positive emotions (pink) occur significantly more often than negative emotions (blue). }
	\label{fig:3}
\end{figure}

\paragraph{Step 3: Quality control.}

Before formal annotation, annotators are asked to annotate training samples until their results pass our examination. 
During the annotation, to eliminate the subjective inconsistency and make the annotation reliable, several specialized workers consistently monitor the collected dialogue data and perform a periodic quality check on samples. After the checking, we sample 10\% data and manually check the samples ourselves. 
If errors are found in an annotation batch, we ask corresponding annotators to self-check and re-annotate the whole batch. 
In light of the above,  the annotation results are closed to real-world natural conversations.

\subsection{Dataset Statistics and Analysis}

The total detailed statistics of the MOD dataset are summarized in Table \ref{tab:1}. MOD dataset has an average of 13.42 turns, and each turn contains 11.46 tokens. The text is tokenized by a Chinese BERT tokenizer \cite{DBLP:conf/nlpcc/WangKZHJZH20}. 
We also plot the usage frequency of Internet memes and corresponding emotion in Figures \ref{fig:2} and \ref{fig:3}, respectively. 
To validate the MOD performance in this work, the final dialogue dataset is divided into training, validation, and testing. 
The split is based on dialogues, not source-target pairs, and the split statistics are summarized in Table \ref{tab:2}. 
In particular, the test set is divided into an easy version for all Internet memes seen in the training set and a hard version for unseen Internet memes. 
The motivation to build a hard version is to evaluate whether a MOD model is able to be transferred to exploit new Internet memes. The phenomenon is common in online chats because a limited candidate cannot handle all cases in real situations. 

\begin{table}[h]
\small
	\begin{center}
		\setlength{\tabcolsep}{1.mm}{
			\begin{tabular}{l|crrr}
				\toprule
				\textbf{Dataset} &\textbf{Type} & \textbf{History} & \textbf{Response} & \textbf{Size}\\
				\midrule
				VisDial \citeyearpar{DBLP:conf/cvpr/DasKGSYMPB17}&task&image+text&text&120K\\
				MMD \citeyearpar{DBLP:conf/aaai/SahaKS18}&task&image+text&image+text&150K\\
				AVSD \citeyearpar{DBLP:conf/cvpr/AlAmriCD0CEBMHA19} &task&audio+video+text&text&11K\\
				 SRS \citeyearpar{DBLP:conf/www/GaoCLL0020} &open&text&sticker&340K\\
				 MOD &open&meme+text&meme+text&45K\\
				\bottomrule
		\end{tabular}}
		{\caption{Comparison with other multimodal dialogue datasets.}
			\label{tab:3}}
	\end{center}
\end{table}

\begin{figure*}
	\begin{center}
		\includegraphics[width=1.95\columnwidth]{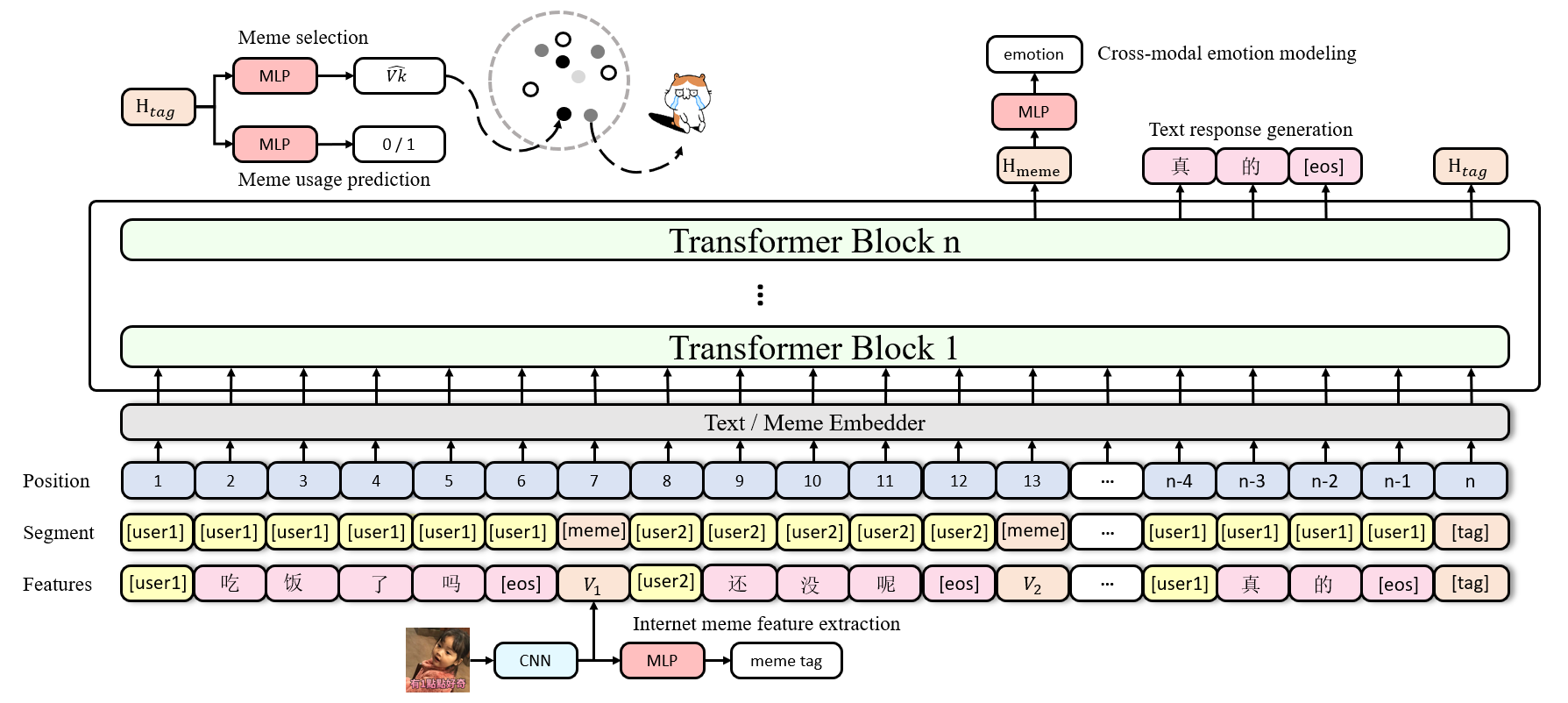}
	\end{center}
	\caption{Overview of MOD-GPT. The proposed model consists of an Internet meme embedder, a text embedder, and a multi-layer Transformer Decoder. The model is optimized with several training tasks.}
	\label{fig:4}
\end{figure*}

To the best of our knowledge, our dataset is novel in the sense that we explicitly guide the annotators to participate in the process of creating engaging and informative MOD.
Table \ref{tab:3} illustrates the comparison between existing multimodal dialogue datasets, and our MOD dataset focuses more on dialogue learning and emotion modeling.  Considering that properly annotated MOD data is still scarce, our dataset will continue to enrich in the future.

\section{MOD-GPT} 

Using the human-annotated dataset, we develop a simple and effective model that aims at incorporating Internet memes into open-domain dialogues. Figure \ref{fig:4} presents the overall multi-task training framework of our model. 

\subsection{Model Overview}

In our method, the response generation is modeled as a language modeling problem. Corresponding, the multimodal response can be produced in order:  
\begin{equation}
    p(u_i|\textbf{U}_{i-1}) = \underbrace{p(m_{i}| {S}_i ,\textbf{U}_{i-1})}_{ meme \ predict} \underbrace{ p(S_i| \textbf{U}_{i-1})}_{text \ generate}
\end{equation}
Practically, we employ a multi-layer Transformer decoder architecture \cite{DBLP:conf/nips/VaswaniSPUJGKP17} to build the probability of output. 

\noindent
$\bullet$  \textbf{Text Input}.
For texts, we follow CDial-GPT \cite{DBLP:conf/nlpcc/WangKZHJZH20} and tokenize the input sentences with a Chinese BERT tokenizer. 

\noindent
$\bullet$ \textbf{Meme Input}.
For Internet memes, we first use pre-trained EfficientNet \cite{DBLP:conf/icml/TanL19} to extract visual features. Then, all meme features are fed through a fully-connected layer to be projected into the same embedding space as text embedding.

To make our model obtain the ability to distinguish among the different part of input (text, meme, user1, and user2) and make use of the order of sequence, the final representation for each token is obtained by summing up its feature embedding, position embedding, and segment embedding.



\subsection{Mixed Response Generation}

When generating Internet meme incorporated response, there exist three problems: ($i$) what text context to respond (text generation),  ($ii$) whether to utilize an Internet meme at current respond (binary classification) and ($iii$) which Internet meme to be selected (meme selection).

For the first text response modeling, we formulate with conventional conditional probability and optimizes the negative log-likelihood loss function on training dataset $D$ as:
\begin{equation}
\small
    \mathcal{L}_{TR} = -E_{\textbf{U} \sim D} [\text{ log} \prod_{l=1}^{L}p(w_{l,i}| w_{<l,i}, \textbf{U}_{i-1})]
\end{equation}
where $w_{l,i}$ is the $l$-th word in $i$-th text response and $L$ corresponds to the length of sentence.

We address the latter two problems in a uniform way in this work. Suppose the original token vocabulary is $V$, we extend it with an extra token “[tag]”, which will be inserted after token “[eos]” at each response. 
During the inference, the model can decide to either retrieve an Internet meme to be inserted from the set or not to use Internet memes.
By utilizing this special token, the questions are merged. For the meme usage prediction, we utilize a one-layer MLP $f$ to project hidden states $\textbf{H}_{tag}$ into two type distribution and optimize with cross-entropy loss as:
\begin{equation}
\begin{split}
        \mathcal{L}_{UP} =& -E_{\textbf{U} \sim D}  [ y \text{ log }f(\textbf{H}_{tag}) \\ &+ (1-y) \text{log} (1 - f(\textbf{H}_{tag}))]
\end{split}
\end{equation}
where for each utterance $u_i$, if $m_i$ = None, then $y = 0$; else $y = 1$.
For predicting meme features, unlike textual tokens, which are represented as discrete labels, meme features are high-dimensional and continuous. Instead of clustering meme features to discrete labels, we adopt the meme feature regression method following \cite{DBLP:journals/corr/abs-2002-00163}. In particular, we apply another one-layer MLP $g$ to transform the hidden states $\textbf{H}_{tag}$ to a vector of the same dimension as target meme feature $v_{correct}$ and optimize the model with the L$2$ loss as: 
\begin{equation}
       \mathcal{L}_{MS} =   E_{\textbf{U} \sim D} \ ||g(\textbf{H}_{tag}) - v_{correct}||_2^2
\end{equation}

\subsection{Learning Objective} 
For meme usage prediction and meme selection, both auxiliary tasks are trained with the main text generation loss together, which can be regarded as the multi-task learning task. The loss function of multi-task learning consists of text response loss $\mathcal{L}_{TR}$, meme usage prediction loss $\mathcal{L}_{UP}$ and meme selection loss $\mathcal{L}_{MS}$. In this way, the total loss function of our model can be computed as follows:
\begin{equation}
    \mathcal{L} = \mathcal{L}_{TR} + \lambda_1 \mathcal{L}_{UP} + \lambda_2 \mathcal{L}_{MS}
\end{equation}
where $\lambda_1$ and $\lambda_2$ are hyper-parameters that work as scale factors.

\subsection{Pre-training Tasks}
\paragraph{Internet Meme Feature Extraction.}

Existing convolutional neural networks (CNN), including EfficientNet, are mostly built on real-world photos. Thus, directly applying these networks on Internet memes to extract features is not feasible. In the dataset \cite{DBLP:conf/www/GaoCLL0020}, each sticker is given an emoji tag, which denotes its general emotion. Considering the relevance between Internet memes and stickers, we adopt a pre-training classification task to help the model understand memes effectively. In particular, we utilize the features output from the CNN to predict which emoji is attached to the corresponding sticker.  An extra MLP layer is integrated, and the cross-entropy loss is used as the optimization function. 

\paragraph{Cross-modal Emotion Modeling.}
The initial parameters of our MOD-GPT model are loaded from the model trained on the text-only Chinese corpus, which lacks cross-modal knowledge building. Thus, we utilize extra emotion labels contained in our dataset and introduce an emotion analysis to help model better handle Internet meme content. Specifically, given the conversation history, the system aims to predict the emotion labels when utilizing the Internet meme in the last utterance, which can also be regarded as a classification problem.  Note that we re-sample top-100 emotion annotations to avoid the training bias.


\section{Experiments} 
\subsection{Baseline Settings}

As discussed earlier, the MOD task has under-investigated so far, and there are few existing baselines for our comparison. Because of the relevance with our MOD task, we select to compare with the following models: 
\begin{itemize} 
    \item SRS (text history and meme response) \cite{DBLP:conf/www/GaoCLL0020} learns the representation of each utterance using a self-attention mechanism and extracts meme representation by CNN. A deep interaction network is employed to fully model the dependency between the sticker and utterances to obtain the target meme.
    \item CDial-GPT (text history and text response)  \cite{DBLP:conf/nlpcc/WangKZHJZH20} is a 12-layer Transformer Decoder under the setting of DialoGPT for Chinese conversation generation.   
    \item MHERAD \cite{DBLP:conf/aaai/SahaKS18} is a multimodal hierarchical encoder-decoder model that incorporates the visual features into the basic HRED model and achieves a promising performance for task-oriented dialogue in the retailing domain.
\end{itemize}


\subsection{Implementation Details}
Since our dataset is multi-turn, we take every sentence in the dialogue from the second sentence to the last sentence as the response of dialogue history.
We implemented the MOD-GPT model with Pytorch under the HuggingFace framework \cite{DBLP:journals/corr/abs-1910-03771}.  The Transformer configure parameter is identical to the base version of CDial-GPT.
ADAM \cite{DBLP:journals/corr/KingmaB14} was used to optimize with the initial learning rate of 5$\times$10$^{-5}$. We set both $\lambda_1$ and $\lambda_2$ to 1. The context length is truncated to be 500. The validation set was used for hyper-parameter tuning. All MLP networks used ReLU activation with dropout and one hidden layer. 
Note that for a fair comparison, all the baselines and our model adopt EfficientNet-b0 \cite{DBLP:conf/icml/TanL19} as the basic architecture of meme feature extractor.
All text responses are generated using Nucleus sampling scheme \cite{DBLP:conf/iclr/HoltzmanBDFC20} with a threshold 0.9 and temperature 0.7. 

\subsection{Evaluation Metrics}
For the text response generation, we use perplexity, which measures the language quality of the generated response.
We also report BLEU-2,4 \cite{DBLP:conf/acl/PapineniRWZ02} and distinct $n$-gram \cite{DBLP:conf/naacl/LiGBGD16} scores, which measure the similarity between the generated responses and ground-truth via $n$-gram matching and the number of distinct $n$-gram in generated responses.
For the Internet meme selection, we use R$_{n}$@$k$ as the evaluation metrics where $k$ is set to 1,2,5, and the model prediction is considered to be correct only if the true response is among the top-$k$ entries in the ranked list of $n$ candidates.

\subsection{Quantitative Results}

\paragraph{Evaluating the Text Response.}

The performance of the baselines and MOD-GPT in textual response generation on two test versions are summarized in Table \ref{tab:4}.
We have the following observations: 1) MOD-GPT surpasses the MHERAD regarding all the evaluation scores, demonstrating that the usefulness and superiority of a unified transformer-based structure for text generation compared with hierarchical LSTM. 2) CDial-GPT, which holds the same network architecture while only access to the text history, achieves lower results of evaluation metrics. In contrast, our MOD-GPT benefits from the Internet meme in a historical context to some extent. We attribute this to the rich Internet meme that contains useful information for dialogue systems to infer answers.

\begin{table}[h]
\small
	\begin{center}	\setlength{\tabcolsep}{1.mm}{
		\begin{tabular}{l|crrrr}
			
			\toprule
			&\textbf{Perplexity}&\textbf{B-2}&\textbf{B-4}&\textbf{Dist-1}&\textbf{Dist-2}\\
		
			\midrule
				\multicolumn{5}{l}{\emph{easy test}}\\ \midrule 
				MHERAD&31.50&2.10&0.46&1.32&13.75\\
			CDial-GPT&19.32&5.63&1.15&1.53&18.62\\
			MOD-GPT&\textbf{19.18}&\textbf{6.06}&\textbf{1.88}&\textbf{1.61}&\textbf{19.25}\\ 
			\midrule
		    \multicolumn{5}{l}{\emph{hard test}}\\ \midrule 
							MHERAD&32.27&2.05&0.44&1.40&14.43\\
			CDial-GPT&19.68&5.54&1.18&1.68&20.22\\
			MOD-GPT&\textbf{19.41}&\textbf{5.98}&\textbf{1.76}&\textbf{1.80}&\textbf{21.65}\\ 
			\bottomrule
		\end{tabular} }
	\end{center}
	{\caption{Performance (\%) of the models on the “text response modeling” task on the MOD test set. }
		\label{tab:4}}
\end{table}

\paragraph{Evaluating the Internet Meme Usage.}

For the meme usage prediction task, the binary classification accuracy scores of the two test versions are 89.5\% and 86.2\%, respectively. This shows that MOD-GPT performs well in whether Internet Meme is included in both test sets.


For the meme retrieval task, we only consider those dialogue turns ending in an Internet meme response from the system, and the model is provided with $n$ target Internet meme in which one is relevant, and the others are randomly sampled. Model has to rank the Internet memes in order of their relevance given the multimodal context. 
We report the performance evaluation in Table \ref{tab:6} and obtain the following findings: 1) MOD-GPT achieves the best performance in this task. In particular, the R$_{10}@5$ score of MOD-GPT approaches to 0.83. 
In contrast, SRS only calculates the similarity between the textual features of the chatting context and the visual features of memes, which results in a degraded performance. 2) As we expected, the unified Transformer-based model holds better retrieval capability compared with the hierarchical LSTM-based model MHERAD. 3) The performance of the hard version of test set, especially for the unseen memes, is worse than the corresponding easy version, reveals that the hard version is more challenging and the  generalization ability for various memes is expected. 
4) When candidate size increases up to total meme sets, the recall score decreases a lot, which demonstrates the future promising improvements.

\begin{table}[t] 
\small
	\begin{center}	\setlength{\tabcolsep}{1.5mm}{
		\begin{tabular}{l|cccc}
			\toprule
			&\textbf{R$_{10}$@1}& \textbf{R$_{10}$@2}&\textbf{R$_{10}$@5}&\textbf{R$_{\text{T}}$@5}\\
		
			\midrule
		
				\multicolumn{5}{l}{\emph{easy test}}\\ \midrule 
				MHERAD&44.1&59.6&80.3&25.1\\
			SRS&46.2&61.9&81.1&-\\
			MOD-GPT&\textbf{52.3}&\textbf{64.3}&\textbf{83.6}&\textbf{32.3}\\\midrule
					\multicolumn{5}{l}{\emph{hard test (seen)}}\\ \midrule 
					MHERAD&43.8 &59.0&78.5&23.8\\
			SRS&44.5&60.8&80.4&-\\
			MOD-GPT&\textbf{52.4}&\textbf{63.0}&\textbf{81.7}&\textbf{31.5}\\ \midrule 
			\multicolumn{5}{l}{\emph{hard test (unseen)}}\\ \midrule 
					MHERAD&35.3&43.8&60.1&16.6\\
			SRS&37.0&45.8&63.2&-\\
			MOD-GPT&\textbf{45.4}&\textbf{51.2}&\textbf{70.5}&\textbf{27.0}\\
			\bottomrule
		\end{tabular} }
	\end{center}
	{\caption{Performance (\%) of the models on the “meme retrieval” task in the MOD test set. The evaluation results of the target Internet memes appear in training set (seen) and do not appear in the training set (unseen) are listed separately.}
		\label{tab:6}}
\end{table}

\begin{figure*}
	\begin{center}
		\includegraphics[width=2\columnwidth]{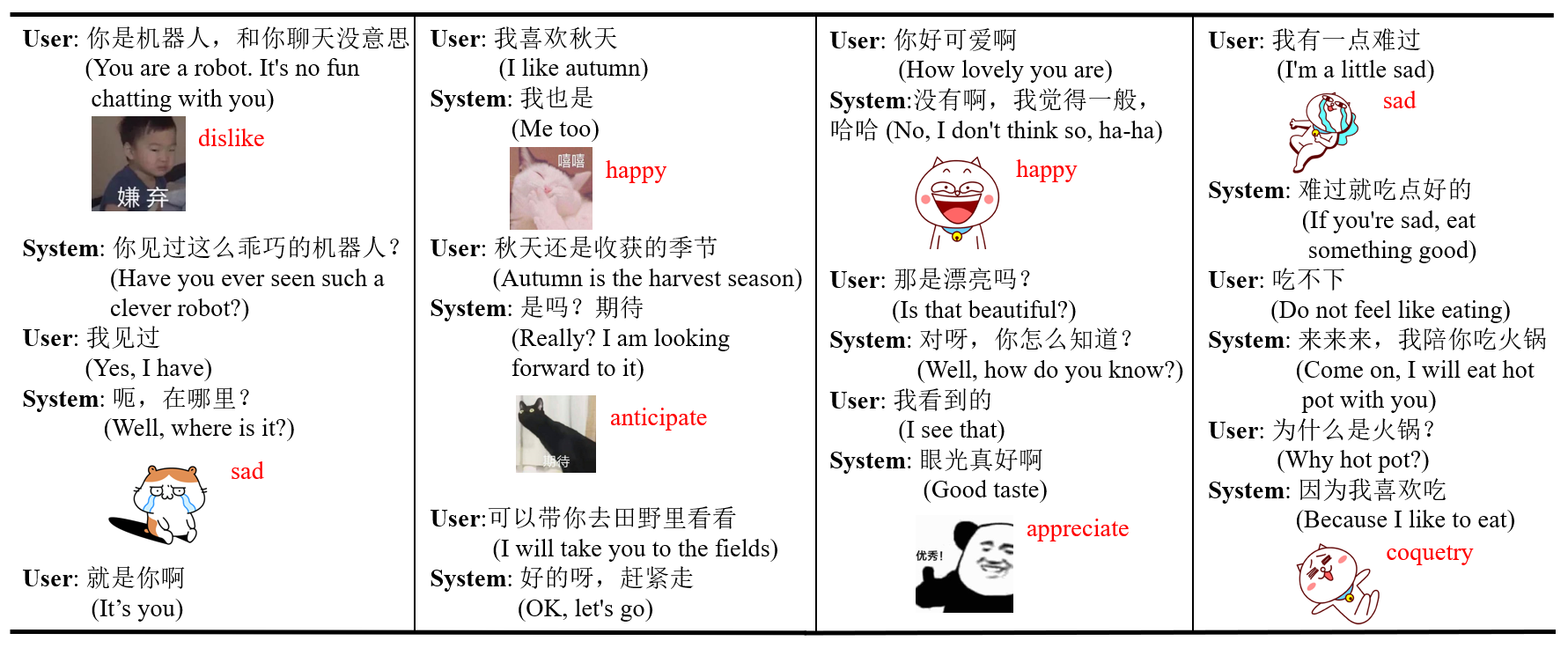}
	\end{center}
	\caption{Examples of Internet meme incorporated dialogues produced by the MOD-GPT (System) and humans (User). All used memes are labeled with corresponding emotions by humans for the convenience of understanding. }
	\label{fig:5}
\end{figure*}

\begin{figure}
	\begin{center}
		\includegraphics[width=0.9\columnwidth]{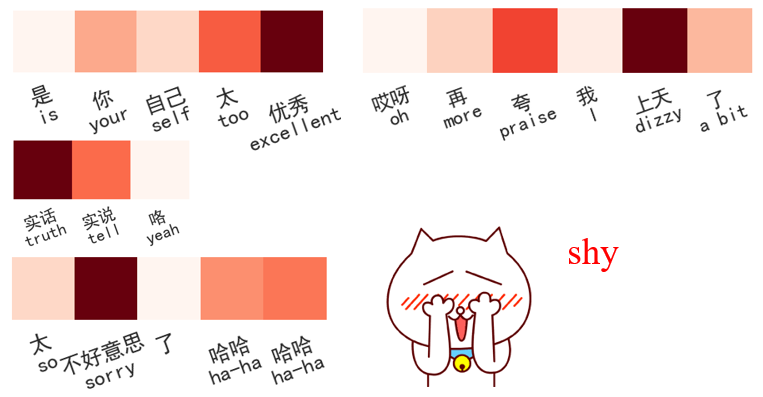}
	\end{center}
	\caption{Examples of the attention weights of the previous dialogue utterance when predicting the target meme with annotated emotion “shy” in red.}
	\label{fig:6}
\end{figure}

\subsection{Case Study}

Several interactive cases generated by MOD-GPT and humans are provided in Figure \ref{fig:5}. These dialogue samples suggest that our proposed method holds the capacity to provide Internet meme incorporated expressive communication.
Besides, according to past studies about Transformer networks \cite{DBLP:conf/emnlp/KovalevaRRR19}, the highest layers of a Transformer model mainly encode task-specific features for predictions. Thus we rely on the mean of all the attention maps on the last layer of the MOD-GPT model to represent the token-level interrelations corresponding to each generated Internet memes, the attention score between the “[tag]” token and each other historical token reflects the token’s contribution to the prediction. We can view that the MOD-GPT model always gives a higher attention weight on the emotional adjectives, such as “sorry” in Figure \ref{fig:6}.

\subsection{Ablation Study}

There are two pre-training tasks in the proposed multimodal dialogue framework. In order to better understand the contribution of each task, we carried out an ablation study for detailed analysis.
As shown in Table \ref{tab:7}, we find that both introduced pre-training tasks help to boost response generation quality partly. Especially, integrating the Internet meme feature extraction (IMFE) training task helps the model better understand Internet memes, which leads to \textbf{+5.1\%} R$_{10}$@1 score improvement compared to original settings. Using cross-modal emotion modeling (CEM) can also bring a remarkable improvement.

\begin{table}[h]
\small
	\begin{center}	\setlength{\tabcolsep}{1.5mm}{
		\begin{tabular}{l|cccc}
			
			\toprule
			&\textbf{B-2}&\textbf{Dist-1}&\textbf{R$_{10}$@1}&\textbf{R$_{10}$@5}\\
			\midrule
		
			MOD-GPT&6.06&1.61&52.3&83.6\\ 
			-IMFE&5.95 &1.35&47.2&79.5\\ 
			-CEM&5.82 &1.27&48.4&80.2\\ 
			\bottomrule
		\end{tabular} }
	\end{center}
	{\caption{Performance (\%) of ablation study on the MOD easy test set. Both pre-training tasks have an impact on improving performance.}
		\label{tab:7}}
\end{table}

\section{Conclusion}

As Internet memes widely propagate through social networks, 
an interesting research direction for future work is incorporating Internet memes into the open-domain dialogue to make conversations vivid and engaging. In this paper, we take a step in this direction by: ($i$) the release of the MOD dataset that focuses on a suitable form of responses according to multimodal historical context with abundant emotion labeling,  ($ii$) a demonstration of a neural conversation system that can generate Internet meme incorporated dialogues under a simple and effective framework. We believe that the ability to deal with the MOD task can serve as an important testbed for measuring progress toward multimodal open-domain dialogue intelligence.

\bibliography{anthology,custom}
\bibliographystyle{acl_natbib}

\clearpage 

\appendix

\section{Appendix}
\label{sec:appendix}

\subsection{Ethical Considerations} 
The original copyright of all the conversations belongs to the source owner that is public to academic use. 
The Internet meme sets are freely accessible online. The copyright of annotation belongs to our group, and they will be free released to the public. 
By consulting the legal advisor, the MOD dataset is freely accessible online to academic use. Without permission, it may not be used for any commercial purposes and distributed to others. 

Our data construction involves manual annotation. The annotated  conversation corpus and Internet meme set do not contain personal sensitive information. 
We asked annotators to incorporate Internet memes limited to given dialogues, and not to include any personal information. 
The annotators got reasonable salary for their annotation work.

\subsection{Technical Difference with Other Multimodal Dialogue Models}
Apart from related multimoda dialogue tasks and datasets, we also discuss the technical difference between the preceding multimodal dialogue models. 
 \citet{DBLP:conf/aaai/SahaKS18} introduces a hierarchical structure, which first uses a multimodal encoder to extract the image and text features, and then adopts high-level RNN to model historical dialogue information, which is also referred to as a baseline in our paper. 
\citet{DBLP:conf/sigir/CuiWSHXN19} propose an adaptive decoder, which first determines whether the reply is in the form of text or image before decoding, and then generates the corresponding response. 
In \cite{DBLP:conf/mm/LiaoM0HC18}, a chat session is modeled as a reinforcement learning procedure, and a reward is formed to optimize the answer. 
\citet{DBLP:conf/mm/HeLLCXHY20} further consider the influence of the order of historical information images and text information on answers with a self-attention block. 
Comparatively, we unify the text generation and meme prediction into a long sequence procedure and solve them with a cross-modal GPT-based language model.

\subsection{Emotion Analysis}

To clarify the capability of our MOD-GPT model in emotion prediction, we illustrate the performance over top-5 classification accuracy metrics in MOD test set in Table 7. As show in the Table, we can see that our method can predict the emotion states of Internet meme usage effectively. This proves that it is reasonable to exploit the conversation emotion for boosting Meme incorporated open-domain dialogue modeling. 

\begin{table}[t] 
\small
	\begin{center}	\setlength{\tabcolsep}{3mm}{
		\begin{tabular}{l|c}
			\toprule
			&\textbf{Emotion Classification Accuracy}\\
			\midrule
			easy test&75.8 \\
			hard test&72.3\\
			\bottomrule
		\end{tabular} }
	\end{center}
	{\caption{Performance (\%) of the MOD-GPT on the “emotion analysis” task in the MOD test set.}
		\label{tab:7}}
\end{table}
\end{document}